\documentclass[10pt,twocolumn,letterpaper]{article}

\usepackage{wacv}                         
\usepackage[accsupp]{axessibility} 

\usepackage{multirow}        
\usepackage{colortbl}        
\newcommand{\oursrow}{\rowcolor{gray!10}[\dimexpr\tabcolsep-\arrayrulewidth\relax][\dimexpr\tabcolsep-\arrayrulewidth\relax]}
\usepackage{threeparttable}  
\usepackage{algorithm}       
\usepackage{algorithmic}     
\usepackage{amsfonts}
\usepackage{tikz}            
\usetikzlibrary{shadows,positioning,calc,arrows.meta}

\newcommand{\sys}{\textsc{EGDR}\xspace}

\newcommand{\vvis}{\mathbf{v}}

\newcommand{\taueff}{\tau_{\text{eff}}}
\newcommand{\Hlo}{H_{\text{lo}}}
\newcommand{\Hhi}{H_{\text{hi}}}
\newcommand{\escaled}{\hat{e}}

\DeclareMathOperator{\clip}{clip}

\definecolor{wacvblue}{rgb}{0.21,0.49,0.74}
\usepackage[pagebackref,breaklinks,colorlinks,allcolors=wacvblue]{hyperref}
\hypersetup{
  pdftitle={Entropy-Gated Dynamic Rectification for Object Hallucination in Large Vision-Language Models},
  pdfauthor={Riasad Alvi, Md. Asaduzzaman Shuvo, Md. Ahashan Habib, Md. Faiyaz Abdullah Sayeedi}
}

\def\wacvPaperID{3757}
\def\confName{WACV}
\def\confYear{2027}

\title{Entropy-Gated Dynamic Rectification for Object \\ Hallucination
in Large Vision-Language Models}

\author{Riasad Alvi$^{1}$, Md. Asaduzzaman Shuvo$^{1}$, Md. Ahashan Habib$^{1}$, Md. Faiyaz Abdullah Sayeedi$^{1,2}$\\
$^{1}$United International University, Bangladesh \quad
$^{2}$BRAC University, Bangladesh\\
{\tt\small \{ralvi212069, ashuvo221104, mhabib201261, msayeedi212049\}@bscse.uiu.ac.bd}
}

\begin{document}
\maketitle

\begin{abstract}
Large Vision-Language Models (LVLMs) routinely name objects that are absent from
the image. Latent steering can reduce this failure in one model forward pass,
but a fixed gate treats early and late decoding steps by the same criterion.
We observe that predictive entropy varies sharply within free-running
generation, while states that commit to short answers occupy a narrow,
confident band. We exploit this temporal mismatch by using uncertainty as a
causally available, token-local control signal for an existing steering gate. We propose
\textbf{E}ntropy-\textbf{G}ated
\textbf{D}ynamic \textbf{R}ectification (EGDR), which lowers the intervention
threshold when entropy derived from the preceding step's already available
logits is high and retains the calibrated operating point when it is low. On
LLaVA-1.5-7B, the underlying static gate lowers CHAIR $C_S/C_I$ from
$53.00/15.30$ to $32.00/14.30$; replacing only its constant threshold with
EGDR further lowers them to $29.00/13.60$. At a matched $58\%$ intervention
rate, the corresponding static control obtains $30.50/14.00$. The gain over the static gate
holds on LLaVA-NeXT-7B and grows with the generation budget, while POPE
scores and general multimodal capability remain effectively unchanged.
EGDR requires no additional model forward pass and adds roughly $2\%$
measured latency. Our code is available at
\mbox{\url{https://github.com/alvi-uiu/EGDR}}.
\end{abstract}
\section{Introduction}
\label{sec:intro}

\begin{figure}[t]
    \centering
    \input{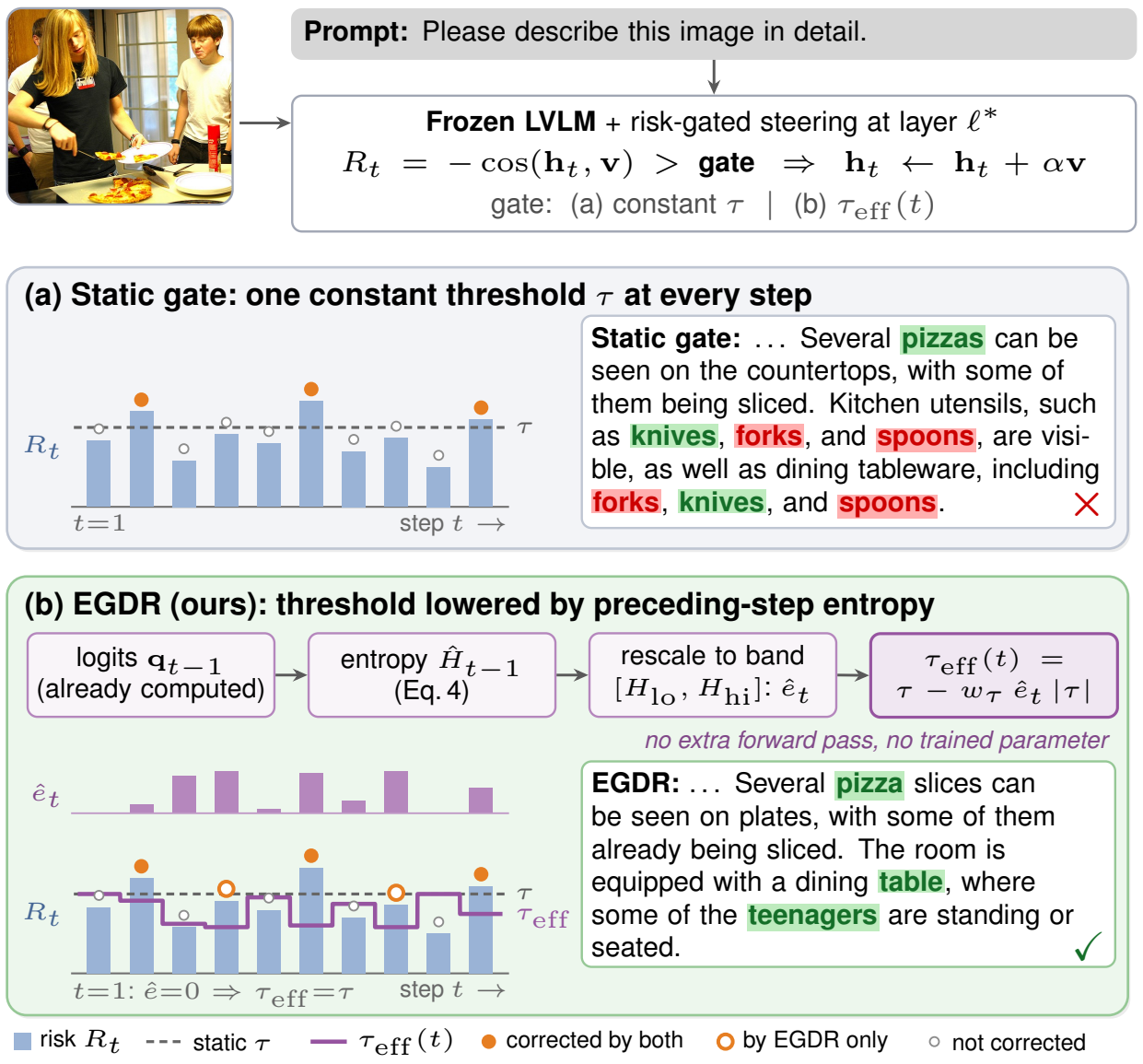}
    \caption{\textbf{Overview of \sys.} Both gates share $\vvis$, $\ell^*$ and
    $\alpha$; only the threshold differs. \textbf{(a)}~A static gate applies one
    threshold $\tau$ at every step. \textbf{(b)}~\sys{} lowers it to $\taueff(t)$
    using the entropy of the preceding step's logits, rescaled to the calibrated
    band, and so corrects states the static gate misses; at $t{=}1$,
    $\escaled{=}0$ and $\taueff{=}\tau$. Gate strips are schematic; bubbles quote
    the second paragraph of real LLaVA-1.5-7B captions verbatim (CHAIR:
    hallucinated objects in red, grounded in green).}
    \label{fig:overview}
\end{figure}

\begin{figure*}[t]
    \centering
    \includegraphics[width=0.99\linewidth]{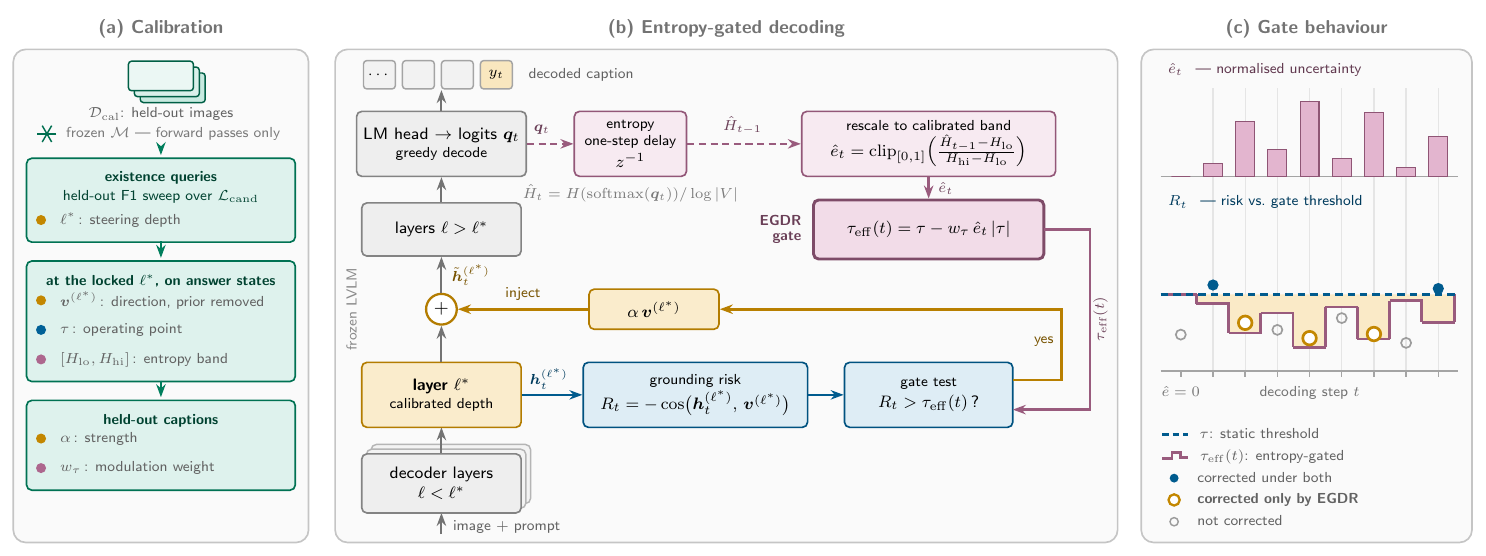}
    \caption{Pipeline of \sys. \textbf{(a)}~Held-out calibration selects the
    steering depth $\ell^{*}$, estimates the prior-removed direction
    $\vvis^{(\ell^{*})}$, threshold $\tau$, and entropy band $[\Hlo,\Hhi]$, and
    tunes $\alpha$ and $w_\tau$. \textbf{(b)}~At each decoding step, the current
    hidden state yields risk $R_t$, while preceding-step entropy lowers $\tau$
    to $\taueff$. If $R_t>\taueff$, \sys{} injects
    $\alpha\vvis^{(\ell^{*})}$. This requires no additional forward pass or
    trained parameter. \textbf{(c)}~The adaptive threshold admits states missed
    by the static gate; at the first token, $\escaled=0$ and $\taueff=\tau$.}
    \label{fig:design}
\end{figure*}

Large vision-language models (LVLMs) produce fluent descriptions that are
frequently ungrounded: they name objects the image does not contain, a failure
that persists across architectures and is severe enough to block deployment where
precision matters~\cite{rohrbach2018object,li2023evaluating,liu2024survey}.
Broader diagnostics further expose language--vision confusion and
attribute--relation errors~\cite{guan2024hallusionbench,wang2024amber}. The cause
is commonly traced to the language backbone asserting its own priors when the
visual evidence is weak or ambiguous.

Training-time curation and robust instruction tuning reduce such failures but
require new data or optimization~\cite{liu2024robust,yu2024hallucidoctor};
post-hoc correction can instead verify and revise an initial response, at the
cost of a multi-stage pipeline~\cite{yin2024woodpecker}. Mitigating hallucination
inside a single decoding pass is therefore attractive. Contrastive decoding compares
the output distribution against one from a degraded or augmented visual
input~\cite{leng2024mitigating,an2025mitigating} or an earlier
layer~\cite{chuang2024dola}; contrasting two inputs is effective but needs
more than one forward pass per token. Latent steering
instead adds a precomputed direction to the hidden state and therefore preserves
the single pass~\cite{li2023inference,liu2024reducing,yang2025nullu,li2025vista}. Because a steering
vector is applied everywhere it would perturb computation that is working
correctly, these methods admit it through a gate: a risk score is computed from
the current hidden state and the correction is injected only when that score
exceeds a threshold $\tau$.

That threshold is a constant. The same value decides whether to intervene at an
early caption token and at the two hundredth. Yet decoder uncertainty is not
constant: the states at which LLaVA-1.5 commits to short answers occupy a narrow
normalized-entropy band between $0.008$ and $0.059$, while the majority of
free-running generation steps lie above it. This exposes a mismatch between a
time-varying decoder and a time-invariant gate and suggests uncertainty as a
causally available control signal. We evaluate it against both the original static gate and
a firing-rate-matched alternative.

We use that signal to set the gate. At each step we read the entropy of the
distribution produced at the previous step, rescale it to a band estimated during
calibration, and lower the threshold in proportion:
$\taueff = \tau - w_\tau \hat{e}_t |\tau|$. Confident steps keep the calibrated
operating point; uncertain steps are corrected more readily. The logits
already exist; converting them to entropy adds a lightweight
vocabulary reduction but no model forward pass and no trained parameter. We call
it \sys, Entropy-Gated Dynamic
Rectification; Fig.~\ref{fig:overview} illustrates it and Fig.~\ref{fig:design}
details the pipeline.

The design follows from two further observations. The depth at which steering
is effective is narrow and non-monotonic: F1 varies by $16.24$ points across five
adjacent layers and peaks at the penultimate one, so we select the layer by
calibration rather than by convention. And the mechanism is confined to
multi-token generation, since a causally available entropy estimate does not exist at the
first decoded token: on discriminative benchmarks answered in one token the
effective threshold reduces to $\tau$ and \sys{} uses the same decision-time
intervention as static steering. This delimits where the
method applies and explains why its benefit appears in captioning while
discriminative accuracy is preserved.

On LLaVA-1.5-7B, a calibrated static gate lowers CHAIR$_I$ from $15.30$ to
$14.30$ and CHAIR$_S$ from $53.00$ to $32.00$. Replacing only its constant
threshold with \sys{} further lowers the two metrics to $13.60$ and $29.00$.
A static threshold matched to \sys{} in firing rate reaches $14.00$ and $30.50$,
respectively. \sys{} also improves on the static gate on LLaVA-NeXT-7B, while POPE, MME and
MM-Vet remain effectively unchanged and measured overhead is approximately
$2\%$ per token.

Our contributions are:
\begin{itemize}\itemsep2pt
\item We identify a temporal mismatch between a fixed steering threshold and
      non-stationary predictive uncertainty, and test preceding-step entropy as
      a causally available control signal without adding a model forward pass.
\item We introduce \sys, modulating the gate rather than the steering
      direction or its magnitude, and show that its benefit is not explained by
      intervening more often: a static gate matched in firing rate recovers
      only part of it.
\item We show that the benefit grows monotonically with the generation budget
      and vanishes on single-token answers: this is the behavior the mechanism
      predicts, and one a discriminative benchmark does not expose.
\item We use held-out validation, rather than an architectural convention, to
      select intervention depth and strength, and compare all gates at that
      locked operating point.
\end{itemize}
\section{Related work}
\label{sec:related}

\subsection{Object hallucination in LVLMs}

Contemporary LVLMs attach a visual encoder to a language backbone through a
projection module, as in BLIP-2~\cite{li2023blip2}, MiniGPT-4~\cite{zhu2023minigpt4},
InstructBLIP~\cite{dai2023instructblip}, Qwen-VL~\cite{bai2023qwenvl} and the
LLaVA family~\cite{liu2023visual,liu2024improved}, which pairs a
CLIP~\cite{radford2021learning} encoder with an MLP connector. Cross-attention
and grounded-token variants include Flamingo, Kosmos-2 and
DeepSeek-VL~\cite{alayrac2022flamingo,peng2023kosmos2,lu2024deepseek}.
Across these designs the model may describe objects, attributes or relations
that the image does not contain~\cite{rohrbach2018object,liu2024survey}. The two standard
probes measure this from opposite directions: CHAIR~\cite{rohrbach2018object}
counts hallucinated objects inside free-form captions, while
POPE~\cite{li2023evaluating} poses balanced yes/no existence questions and is
therefore answered in a single token; HallusionBench and AMBER broaden diagnosis
to language--vision confusion, attributes and relations~\cite{guan2024hallusionbench,wang2024amber}.
The generative--discriminative distinction matters for any method
whose signal is defined over a sequence of decoding steps, and we return to it in
Sec.~\ref{sec:analysis_scope}.

\subsection{Inference-time mitigation}

Robust instruction tuning and cleaning hallucinatory visual instructions improve
the model before deployment~\cite{liu2024robust,yu2024hallucidoctor}, while
Woodpecker verifies and revises a generated response through a multi-stage
post-hoc pipeline~\cite{yin2024woodpecker}. These approaches require data,
optimization or additional modules, so a large body of work intervenes directly
at inference. One family manipulates output
distributions. VCD~\cite{leng2024mitigating} contrasts the logits obtained from
an image against those from a distorted copy, AGLA~\cite{an2025mitigating}
fuses the original-image logits with those of a prompt-relevant augmented view,
DoLa~\cite{chuang2024dola} contrasts early-exit and final-layer predictions,
OPERA~\cite{huang2024opera} penalizes the over-trust pattern that precedes
hallucinated tokens during beam search, and M3ID~\cite{favero2024multi} contrasts
the conditioned and unconditioned distributions. Contrasting two inputs costs a
second forward pass per token and OPERA adds beam search; DoLa reads earlier
layers of the same pass. Several authors have noted that contrasting away language priors can also
remove priors the model legitimately needs~\cite{chen2025ict}.

Other training-free decoders exploit internal structure: DeCo mixes evidence
from selected preceding layers~\cite{wang2025deco}, DAMRO suppresses visually
irrelevant high-attention tokens~\cite{gong2024damro}, and FarSight modifies the
causal attention mask to limit snowballing errors~\cite{tang2025farsight}. These
methods adapt logits or attention; \sys{} instead adapts whether a fixed latent
correction is admitted.

A second family edits internal state directly. PAI~\cite{liu2024paying}
rebalances attention toward image tokens and refines the logits against a
text-only input, ICT~\cite{chen2025ict} shifts the activations of
attention heads that carry image-level and object-level evidence, and
ONLY~\cite{wan2025only} reports that intervening at a single layer in a single
query is sufficient: the design point we adopt. Attention-lens and head-suppression analyses likewise
localize hallucination to particular middle-layer pathways or dynamically
selected heads~\cite{jiang2025devils,sarkar2025spin}.

\subsection{Latent steering}

Closest to our setting are methods that add a precomputed direction to the hidden
state at inference. The idea originates with inference-time
intervention~\cite{li2023inference} for language models and has been adapted to
the multimodal case by VTI~\cite{liu2024reducing}, steering latent
representations along a direction estimated from contrastive pairs, by VISTA,
which combines visual-information steering with early-layer token evidence~\cite{li2025vista}, and by
Nullu~\cite{yang2025nullu}, projecting representations out of a subspace
associated with hallucination. To isolate temporal gating, we instantiate a
risk-gated steering baseline: a risk score is computed from the current hidden
state and the direction is injected only when that score exceeds a fixed
threshold. The visual direction used in our experiments is
constructed from the model itself, by contrasting the hidden states produced
with and without visual input and removing the component aligned with the
language-prior direction (Sec.~\ref{sec:analysis_setup});
Sec.~\ref{sec:analysis} examines what such a threshold should depend on, and
Sec.~\ref{sec:method} derives it from the model's own uncertainty.

This family leaves open \emph{where} and \emph{when} to intervene. Steering
applied uniformly across layers disturbs unrelated computation and costs general
capability~\cite{yang2025nullu,liu2024reducing}; we select a single depth by
calibration (Sec.~\ref{sec:analysis_depth}). A fixed gate also treats all
decoding steps as equally at risk, although errors accumulate autoregressively.
We address this second point. Distribution-level decoders (VCD, DoLa, OPERA,
M3ID) reshape the output distribution, mostly with extra forward passes;
\sys{} neither
contrasts distributions nor alters the steering vector, and uses preceding-step
entropy only to decide \emph{when} a fixed latent correction is admitted.

The static arm reported throughout our experiments is exactly this
formulation: a direction projected onto the orthogonal complement of the
language prior, a single intervention depth fixed by calibration, and a
threshold placed at a percentile of the risk observed on factual states;
\sys{} recovers it exactly at $w_\tau=0$. Every difference we report between
the two arms is therefore the gate and nothing else.
Concurrent work intervenes at other points of the same pipeline: HulluEdit
edits an evidence-consistent subspace in a single pass~\cite{hulluedit2026},
attribution-guided sparsification selects the layers to steer by causal
attribution rather than by calibration~\cite{ltsfs2026}, VEGAS keys an adaptive
intervention on vision-encoder attention rather than on
uncertainty~\cite{vegas2026}, and prefill-time intervention argues that
decoding-stage steering amplifies residual hallucinations because errors
accumulate autoregressively~\cite{pti2026}. Our length sweep in
Sec.~\ref{sec:exp_length} bears directly on that last claim.
Token reduction offers a complementary view of inference-time computation:
compression removes redundant tokens to cut cost~\cite{shao2026survey}, it has
been argued to serve goals beyond efficiency, including hallucination
mitigation~\cite{kong2025token}, and pruning can itself introduce safety
vulnerabilities~\cite{wang2026pruning}. \sys{} removes no token; it uses
uncertainty only to decide at which decoding steps a fixed correction is admitted.

\section{Preliminary analysis}
\label{sec:analysis}

This section asks whether a constant steering threshold is justified. We
establish where the intervention is effective (Sec.~\ref{sec:analysis_depth}),
show that predictive uncertainty during generation lies largely outside the
confident regime of committed answers (Sec.~\ref{sec:analysis_entropy}), and
identify where an uncertainty-dependent threshold can act
(Sec.~\ref{sec:analysis_scope}). All measurements use LLaVA-1.5-7B with greedy
decoding on calibration data disjoint from every evaluation set.

\subsection{Risk-gated steering}
\label{sec:analysis_setup}

We use a risk-gated steering formulation. Let $\mathbf{h}_t^{(\ell)}$ be the
hidden state at layer $\ell$ and decoding step $t$. For each correctly answered
calibration query, we replay the same answer with the image masked and average
the paired state difference to obtain $\mathbf{v}_{\rm raw}^{(\ell)}$. We also
form a language-prior direction $\mathbf{u}_{\rm prior}^{(\ell)}$ from the mean
difference between incorrect and correct answer states when the image is
masked. All states are taken at the last answer token. The fixed direction used
by every gated arm is
\begin{equation}
 \mathbf{v}^{(\ell)}=\operatorname{norm}\!\left(
 \mathbf{v}_{\rm raw}^{(\ell)}-
 \langle\mathbf{v}_{\rm raw}^{(\ell)},\mathbf{u}_{\rm prior}^{(\ell)}\rangle
 \mathbf{u}_{\rm prior}^{(\ell)}\right),
 \label{eq:direction}
\end{equation}
where $\mathbf{u}_{\rm prior}^{(\ell)}$ is unit-normalized and
$\operatorname{norm}$ denotes unit normalization. The raw direction has cosine
$0.40$ with the prior direction; the projection reduces it below $0.02$ and
improves CHAIR under an otherwise identical configuration (supplementary
material). This construction is held fixed in all comparisons, so the claimed
contribution is the gate rather than a new steering direction.
The hallucination risk of a state is its negative alignment with that direction,
\begin{equation}
    R(\mathbf{h}) = - \cos\!\big(\mathbf{h}, \mathbf{v}^{(\ell)}\big),
    \label{eq:risk}
\end{equation}
and steering fires when the risk exceeds a threshold $\tau$:
\begin{equation}
    \tilde{\mathbf{h}}_t^{(\ell)} = \mathbf{h}_t^{(\ell)}
    + \alpha \cdot \mathbf{1}\big(R_t > \tau\big) \cdot \mathbf{v}^{(\ell)} .
    \label{eq:static}
\end{equation}
This mechanism is governed by the depth $\ell$ at which it acts, the strength
$\alpha$, and the threshold $\tau$. We select depth and strength on held-out
calibration data before comparing gate designs.

\subsection{Steering depth is not a free parameter}
\label{sec:analysis_depth}

\textbf{Observation.} We apply Eq.~\ref{eq:static} at decoder layers $27$--$31$,
holding $\alpha$ and the threshold calibration fixed, and evaluate each choice
on held-out $500$-question development panels. F1 is non-monotonic: it rises
from $82.10$ at layer $27$ to $88.64$ at layer $30$, then falls to $72.40$ at
layer $31$ (supplementary material).

\textbf{Implication.} A representation-space separability heuristic is not a
reliable depth selector: the final layer may separate factual and erroneous
states yet leave no downstream block in which an edit can propagate. We
therefore define depth selection operationally as held-out task performance,
$\ell^*=\arg\max_{\ell\in\{27,\ldots,31\}}\mathrm{F1}_{\rm dev}(\ell)$, which
selects layer $30$ for LLaVA-1.5-7B. The candidate set is fixed before the gate
ablation and the same selection rule is rerun for the second backbone.

\textbf{Strength is calibrated generatively.} The gate targets multi-token
generation, so we select $\alpha$ on held-out captions rather than on the
development queries.

\subsection{\texorpdfstring{Predictive uncertainty varies\\within a generation}{Predictive uncertainty varies within a generation}}
\label{sec:analysis_entropy}

\textbf{Observation.} A constant threshold does not use the substantial temporal
variation in the decoder's confidence. We quantify that variation with the
normalized predictive entropy
\begin{equation}
    \hat{H}_t = \frac{H\big(p(\cdot \mid x_{<t})\big)}{\log |V|} \in [0,1]
    \label{eq:entropy}
\end{equation}
at every decoding step. The quantity is strongly non-stationary. The states at
which the model commits to an answer occupy a narrow, highly confident band:
writing $\Hlo$ and $\Hhi$ for its tenth and ninetieth percentiles, we measure
$\Hlo = 0.008$ and $\Hhi = 0.059$,
whereas the majority of free-running generation steps lie above it: $62.5\%$ of
steps on binary existence questions and $71.6\%$ on $512$-token captioning
exceed the ninetieth percentile of committed states.

\textbf{Implication.} These distributions establish the temporal signal needed
by an adaptive scheduler. We use preceding-step entropy as the control variable
and evaluate it against both the original static gate and a static gate matched
in intervention rate. Computing entropy uses existing logits and adds no model
forward pass; its measured cost is reported in Sec.~\ref{sec:exp_cost}.

\subsection{The regime in which an adaptive gate can act}
\label{sec:analysis_scope}

\textbf{Observation.} A causally available estimate of $\hat{H}_t$ uses the distribution
produced at step $t-1$; at the first generated token no such distribution exists,
so $\hat{H}_0 = 0$ by construction and any entropy-dependent threshold reduces
exactly to its static value, so the first-token decision on a binary existence
question is the same under both gates. A sweep over the modulation weight on the development panels confirms it: the
firing rate moves from $20.8\%$ to $75.8\%$ while first-token accuracy and F1
are identical (supplementary material). POPE labels, however, are parsed
from the generated answer, whose later tokens the gate can affect.

\textbf{Implication.} The gate is a mechanism for open-ended generation, where it
has many decoding steps on which to act. On single-token answers it reduces to
the calibrated static threshold, so it can be enabled by default without altering
first-token decisions. We evaluate the gate on descriptive captioning and use
the discriminative benchmark to confirm that this behavior is preserved.

\noindent\textbf{Summary.} Validation performance selects the intervention
depth, while strongly non-stationary predictive uncertainty supplies the control
signal for the adaptive gate. The matched controls in Sec.~\ref{sec:results}
isolate entropy-conditioned timing from a uniformly lower threshold.

\section{\texorpdfstring{\sys: Entropy-Gated\\Dynamic Rectification}{EGDR: Entropy-Gated Dynamic Rectification}}
\label{sec:method}

Sec.~\ref{sec:analysis} established that intervention depth must be selected on
held-out performance and that decoder uncertainty is non-stationary. \sys{}
tests whether the latter can improve gate timing: it makes
the gate, rather than the injected direction, a function of predictive entropy.
The method is calibration-based rather than trained: it adds no parameters or
gradient updates and no additional forward pass, while its depth, threshold,
entropy band, strength and modulation weight are set by forward passes on a
held-out pool ($100$ images and $500$ queries per replicate, plus held-out
captions;
Sec.~\ref{sec:exp_setup}). It proceeds in three parts: a calibrated threshold that fixes the
operating point (Sec.~\ref{sec:method_tau}), an entropy-dependent modulation of
that threshold (Sec.~\ref{sec:method_gate}), and held-out selection of the
intervention depth (Sec.~\ref{sec:method_depth}); Fig.~\ref{fig:design}
summarizes the three parts.

\subsection{Calibrated operating point}
\label{sec:method_tau}

The risk score of Eq.~\ref{eq:risk} is a cosine similarity, so its scale depends
on the layer and on the model. A threshold expressed as an absolute constant does
not transfer between them. We therefore define the threshold as a quantile of the
risk distribution observed on states that the model produces when it answers
correctly. Given a calibration set $\mathcal{D}_{\text{cal}}$ of images with
binary existence queries, we collect the risk scores of factual states,
$\mathcal{S}_{\text{fact}} = \{R(\mathbf{h}) \mid \mathbf{h} \in
\mathcal{H}_{\text{fact}}^{(\ell^{*})}\}$, and set
\begin{equation}
    \tau = \operatorname{Percentile}\big(\mathcal{S}_{\text{fact}},\, k\big).
    \label{eq:tau}
\end{equation}
The gate then opens only when a state is less visually grounded than the
$k$-quantile of the model's own correct behavior. Its nominal factual-state
intervention rate is therefore $1-k$ before distribution shift. On LLaVA-1.5-7B at
$\ell^{*}=30$ with $k=0.8$ this yields $\tau = -0.094$; the negative value is
typical, since deep hidden states are on average weakly anti-aligned with the
visual direction.

\subsection{Entropy-gated threshold}
\label{sec:method_gate}

We now make the threshold a function of the uncertainty measured in
Eq.~\ref{eq:entropy}. The modulation must satisfy two properties: it must be
\emph{causally available}, using only information that exists before the token is emitted, and
\emph{scale-correct}, lowering the threshold when the model is
uncertain irrespective of the sign of $\tau$.

\textbf{Causally available estimate.} At step $t$ we read the entropy of the distribution
produced at step $t-1$. This requires one softmax and reduction over logits that
have already been computed, but no repeat model evaluation. Non-finite logits are
guarded and the estimate is clamped to $[0,1]$.

\textbf{Normalization.} Raw entropy is not comparable across models or tasks: on
the calibration data of Sec.~\ref{sec:analysis_entropy} it occupies a band of
width $0.05$ near zero, so an absolute modulation would be numerically
negligible. We rescale it instead to the range the model actually occupies. Let
$\Hlo$ and $\Hhi$ be the tenth and ninetieth percentiles of entropy at the
answer-decision states in $\mathcal{D}_{\text{cal}}$, and set
\begin{equation}
    \hat{e}_t = \clip_{[0,1]}\!\left(
    \frac{\hat{H}_{t-1} - \Hlo}{\Hhi - \Hlo} \right).
    \label{eq:escale}
\end{equation}
This keeps the modulation strength $w_\tau$ on an $O(1)$ scale. It saturates for
many captioning states; we retain it as the primary configuration because it
uses the same answer-calibration pass as $\tau$, and report a generation-state
normalization only as an ablation.

\textbf{Modulation.} The effective threshold at step $t$ is
\begin{equation}
    \tau_{\text{eff}}(t) = \tau - w_\tau \cdot \hat{e}_t \cdot |\tau| ,
    \qquad w_\tau \ge 0 .
    \label{eq:taueff}
\end{equation}
The multiplication by $|\tau|$ rather than by $\tau$ changes the mechanism, not just its notation. The familiar
form $\tau(1 - w_\tau \hat{e}_t)$ lowers the threshold only when $\tau > 0$; for
the negative thresholds Eq.~\ref{eq:tau} typically produces it raises the
threshold, making an uncertain step \emph{harder} to correct.
Eq.~\ref{eq:taueff} coincides with the familiar form when $\tau > 0$ and remains
correct for either sign.

The steering rule is otherwise unchanged from Eq.~\ref{eq:static}:
\begin{equation}
    \tilde{\mathbf{h}}_t^{(\ell^{*})} = \mathbf{h}_t^{(\ell^{*})}
    + \alpha \cdot \mathbf{1}\big(R_t > \tau_{\text{eff}}(t)\big)
    \cdot \mathbf{v}^{(\ell^{*})} .
    \label{eq:egdr}
\end{equation}
Confident steps retain the calibrated operating point; uncertain steps, at which
the model is selecting among competing continuations, are corrected on a lower
threshold (Fig.~\ref{fig:overview}). The injected magnitude is left constant by design: scaling a vector
that is already being injected changes how strongly the state is moved but not
\emph{when} the intervention occurs, and it is timing that
Sec.~\ref{sec:analysis} proposes as the control variable.

\subsection{Depth calibration}
\label{sec:method_depth}

Sec.~\ref{sec:analysis_depth} showed that representation-space separability does
not identify an effective intervention depth. We fix a small set of late decoder
layers and select
\begin{equation}
    \ell^{*}=\arg\max_{\ell\in\mathcal{L}_{\rm cand}}
    \mathrm{F1}_{\rm dev}(\ell),
    \label{eq:depth}
\end{equation}
using the static gate on held-out existence queries. Only after $\ell^*$ is
locked do we estimate $\tau$, $\Hlo$ and $\Hhi$ at that layer and compare gate
designs. For LLaVA-1.5-7B, $\mathcal{L}_{\rm cand}=\{27,\ldots,31\}$ and the
rule selects layer $30$. The same procedure, rather than the same layer index,
is used for LLaVA-NeXT.

\subsection{Complexity and scope}
\label{sec:method_cost}

\sys{} touches one layer at one point in the forward pass. The risk score is a
single cosine similarity against a cached vector, entropy adds a softmax and
reduction over existing logits, and the injection is one scaled vector addition.
No parameter is trained and no auxiliary model is used. The operation is not
computationally free, but it is small relative to a decoder forward pass;
measured latency is reported in Sec.~\ref{sec:exp_cost}. Contrasting two
inputs, by comparison, requires an additional model evaluation per step.

By Sec.~\ref{sec:analysis_scope}, the mechanism acts only in multi-token
generation: at the first token $\tau_{\text{eff}}=\tau$, so POPE, whose yes/no answer
begins at that token, serves as a preservation check
(Sec.~\ref{sec:exp_pope}) rather than as evidence for the gate.

Algorithm~S1 in the supplementary material summarizes the complete
calibration and decoding procedure.

\begin{table*}[!t]
\setlength{\tabcolsep}{3pt}
\small
\centering
\caption{\textbf{POPE} on LLaVA-1.5-7B (ten-panel mean, $3{,}000$ questions per strategy; \%, higher is better). Both gates share $\tau$ at the first token (Sec.~\ref{sec:analysis_scope}).}
\label{tab:pope}
\setlength{\tabcolsep}{2.2mm}
\begin{tabular}{@{}l|ccc|ccc|ccc}
\toprule
\multirow{2}{*}{\textbf{Method}}
& \multicolumn{3}{c|}{\textbf{POPE -- \textit{Random}} ($\uparrow$)}
& \multicolumn{3}{c|}{\textbf{POPE -- \textit{Popular}} ($\uparrow$)}
& \multicolumn{3}{c}{\textbf{POPE -- \textit{Adversarial}} ($\uparrow$)} \\
& ACC & Rec & F1 & ACC & Rec & F1 & ACC & Rec & F1 \\
\midrule
Regular & 87.75 & 78.69 & 86.53 & 85.95 & 78.69 & 84.85 & 83.25 & 78.69 & 82.45 \\
VCD~\cite{leng2024mitigating} & 86.99 & 79.02 & 85.86 & 84.29 & 79.02 & 83.41 & 81.25 & 78.95 & 80.81 \\
M3ID~\cite{favero2024multi} & 87.69 & 78.42 & 86.43 & 85.92 & 78.42 & 84.78 & 83.22 & 78.22 & 82.34 \\
ONLY~\cite{wan2025only} & 87.05 & 78.62 & 85.86 & 85.09 & 78.62 & 84.05 & 81.85 & 78.55 & 81.23 \\
AGLA~\cite{an2025mitigating} & 87.69 & 78.55 & 86.45 & 85.92 & 78.55 & 84.80 & 83.25 & 78.35 & 82.39 \\
VTI~\cite{liu2024reducing} & 87.59 & 77.49 & 86.19 & 85.25 & 77.49 & 84.01 & 81.79 & 77.29 & 80.93 \\
\midrule
Static gate (matched $\alpha$) & 89.30 & 83.31 & 88.62 & 87.10 & 83.31 & 86.61 & \textbf{83.52} & 83.20 & \textbf{83.51} \\
\oursrow \textbf{\sys{} (ours)} & \textbf{89.32} & \textbf{83.33} & \textbf{88.64} & \textbf{87.11} & \textbf{83.33} & \textbf{86.63} & 83.50 & \textbf{83.22} & 83.49 \\
\bottomrule
\end{tabular}
\end{table*}

\section{Experiments}
\label{sec:results}

We ask whether \sys{} reduces hallucination where its mechanism is active
(Sec.~\ref{sec:exp_chair}), whether the reduction is attributable to the timing
of the intervention rather than to its volume
(Sec.~\ref{sec:exp_length},~\ref{sec:exp_ablation}), and whether behavior
outside that regime is preserved (Sec.~\ref{sec:exp_pope},~\ref{sec:exp_utility}).

\subsection{Setup}
\label{sec:exp_setup}

\textbf{Model and decoding.} Experiments use the frozen LLaVA-1.5-7B
checkpoint (LLaVA-NeXT-7B in Sec.~\ref{sec:exp_next}) in half precision with greedy decoding. No parameter is updated.

\textbf{Benchmarks.} POPE-style queries~\cite{li2023evaluating} use
\textit{random}, \textit{popular} and \textit{adversarial} sampling, a standard
answer-format prompt and a $10$-token budget; labels are parsed from the first
sentence of the answer (a standalone ``no'' or ``not'' yields \textit{no}, as
does an empty answer; otherwise \textit{yes}). CHAIR~\cite{rohrbach2018object} uses COCO val2014
images~\cite{lin2014microsoft}, a detailed-description prompt and a $512$-token
budget; $C_S$ and $C_I$ are the fractions of captions and of object mentions
that are hallucinated. We also report MME~\cite{fu2025mme} and
MM-Vet~\cite{yu2023mm} on their fixed sets.

\textbf{Aggregation.} POPE, CHAIR and development results are means over ten
independently sampled panels, paired across methods ($3{,}000$ balanced
questions per POPE sampling strategy, $500$ CHAIR images, $500$ development
queries); F1 and $C_I$ are computed within each panel. The resampled POPE
panels differ from the official fixed splits. All methods run in our pipeline
on the same panels, prompt, budget and greedy decoding.

\textbf{Implementation and selection.} We steer at $\ell^{*}=30$ with
$\alpha=1.1$, quantile $k=0.8$ and modulation $w_\tau=0.5$. Each calibration
replicate contains $100$ COCO train2014 images and $500$ binary existence
queries, disjoint from every evaluation image; the development panels are such
replicates. Candidate depth is selected by Eq.~\ref{eq:depth};
the direction, threshold and entropy band are then estimated at that locked
layer. Strength and gate settings are selected on held-out captions from the
calibration pool and fixed before any evaluation panel is decoded.
No evaluation image enters a tuning decision. Table~\ref{tab:gate} is a
sensitivity analysis around the locked value, not a test-set selection rule.

\textbf{Matched controls.} To isolate the contribution of the gate we include
two static arms: one identical to \sys{} in direction, depth and
strength and differing only in that its threshold is constant, the
configuration \sys{} reduces to at $w_\tau=0$; the other lowers that constant
threshold until it fires as often as \sys{} does, controlling for the volume
of intervention and leaving only its timing.

\begin{table}[!t]
\small
\centering
\caption{\textbf{CHAIR} on LLaVA-1.5-7B (ten-panel mean, $500$ images, $512$ tokens; lower is better). Len.: mean caption length in words.}
\label{tab:chair}
\begin{tabular}{@{}l|cc|c}
\toprule
\textbf{Method} & $C_S$ ($\downarrow$) & $C_I$ ($\downarrow$) & Len. \\
\midrule
Regular & 53.00 & 15.30 & 96.1 \\
VCD~\cite{leng2024mitigating} & 51.50 & 15.90 & 97.8 \\
M3ID~\cite{favero2024multi} & 49.50 & 14.00 & 94.2 \\
ONLY~\cite{wan2025only} & 50.00 & 14.20 & 93.4 \\
AGLA~\cite{an2025mitigating} & 50.50 & 13.90 & 95.0 \\
VTI~\cite{liu2024reducing} & 36.00 & 15.60 & 78.3 \\
\midrule
Static gate (matched $\alpha$) & 32.00 & 14.30 & 88.4 \\
\oursrow \textbf{\sys{} (ours)} & \textbf{29.00} & \textbf{13.60} & 87.6 \\
\bottomrule
\end{tabular}
\end{table}

\begin{figure*}[t]
    \centering
    \input{fig/f_qual.tex}
    \caption{\textbf{Qualitative comparison of object hallucination mitigation.}
    LLaVA-1.5-7B captions of a COCO val2014 image under unsteered decoding
    (Regular), the calibrated static gate, and \sys{} at the locked operating
    point of Sec.~\ref{sec:exp_setup} (greedy, $512$ tokens). Objects that the
    CHAIR evaluator flags as hallucinated against the COCO annotations are in
    red. Image cropped for display.}
    \label{fig:qual}
\end{figure*}

\subsection{Generative hallucination}
\label{sec:exp_chair}

Table~\ref{tab:chair} reports CHAIR. \sys{} reduces $C_I$ from $15.30$ to
$13.60$ and $C_S$ from $53.00$ to $29.00$. Against the static gate it improves
both metrics, $32.00 \to 29.00$ on $C_S$ and $14.30 \to 13.60$ on $C_I$, and
against the static gate matched in firing rate the metrics improve
again, $30.50 \to 29.00$ and $14.00 \to 13.60$ (Tab.~\ref{tab:gate}). Because
the firing rates are matched, the remaining gap reflects intervention timing
rather than intervention volume.

A caption that names few objects rarely names a wrong one, so $C_S$ can be bought
by describing less. The length column exposes part of this trade-off: VTI reaches $36.00$
$C_S$ with captions of $78.3$ words, an $18\%$ reduction in content, and its
$C_I$ is accordingly no better than Regular. \sys{} shortens captions from
$96.1$ to $87.6$ words while attaining the lowest value of both hallucination
metrics, improving $C_I$ rather than relying on sentence length alone.
Fig.~\ref{fig:qual} shows an example: the unsteered model invents a frisbee,
the static gate still invents backpacks, and \sys{} names neither with a caption
as long as the static gate's.

\subsection{The margin grows with the generation budget}
\label{sec:exp_length}

If the gate acts through the entropy of preceding steps, it can do nothing when
there are none and progressively more as the generation lengthens. Re-running
both arms at three budgets (supplementary material) shows this pattern: at $64$
tokens \sys{} and the static gate are indistinguishable
($\Delta C_S = -0.50$), at $128$ tokens the margin is $-1.50$ and at $512$
tokens $-3.00$, with $\Delta C_I$ moving the same way, $-0.10 \to -0.40 \to
-0.70$. This is the positive counterpart of the null result on POPE: the property
that makes the gate inert on single-token answers makes it strongest on long
captions, where object hallucination is measured.

\begin{table}[!t]
\small
\setlength{\tabcolsep}{3pt}
\centering
\caption{Gate ablation on LLaVA-1.5-7B (ten-panel mean, $512$ tokens). Fire \%: share of decoding steps with the gate open; bold: locked configuration.}
\label{tab:gate}
\begin{tabular}{@{}l|cc|cc@{}}
\toprule
\textbf{Configuration} & $k$ & $w_\tau$ & Fire \% & $C_S$ / $C_I$ ($\downarrow$) \\
\midrule
Static, calibrated $\tau$ & 0.80 & n/a & 21.0 & 32.00 / 14.30 \\
Static, matched firing & 0.45 & n/a & 58.2 & 30.50 / 14.00 \\
\midrule
\sys, $w_\tau{=}0.25$ & 0.80 & 0.25 & 38.1 & 30.50 / 13.90 \\
\oursrow \sys, $w_\tau{=}0.5$ (ours) & 0.80 & 0.50 & 58.0 & \textbf{29.00} / \textbf{13.60} \\
\sys, $w_\tau{=}1.0$ & 0.80 & 1.00 & 71.3 & 29.50 / 13.80 \\
\midrule
\sys, band from generation & 0.80 & 0.50 & 52.4 & 28.50 / 13.50 \\
\bottomrule
\end{tabular}
\end{table}

\subsection{Isolating the gate}
\label{sec:exp_ablation}

Table~\ref{tab:gate} separates the timing of the intervention from its volume.
Lowering the static threshold from $k=0.8$ to $k=0.45$ raises the firing rate
from $21.0\%$ to $58.2\%$, matching \sys{} at $58.0\%$, and recovers part of the
improvement ($32.00 \to 30.50$ on $C_S$) but not the rest: at the same volume of
intervention \sys{} is a further $1.50$ points of $C_S$ and $0.40$ of $C_I$
lower. Because $\escaled_t$ saturates on most caption steps ($71.6\%$ exceed
$\Hhi$, Sec.~\ref{sec:analysis_entropy}), \sys{} largely switches between $\tau$
and a lower threshold; the matched arm tests whether one fixed lower threshold
suffices, and it does not recover the full gain.

The modulation weight is a single interpretable knob: at $w_\tau=0.25$ the gate
fires less often, while at $w_\tau=1.0$ it fires at $71.3\%$ of steps; the
locked $w_\tau=0.5$ setting lies between them. Estimating the entropy band from free-running
generation states rather than from committed-answer states improves both metrics
further ($28.50$ / $13.50$), since the rescaled uncertainty of
Eq.~\ref{eq:escale} then uses its full dynamic range instead of saturating. We
retain the answer-calibrated primary configuration because it estimates
$\tau$, $\Hlo$ and $\Hhi$ in a single calibration workflow.

\subsection{Discriminative behavior}
\label{sec:exp_pope}

Table~\ref{tab:pope} reports POPE. \sys{} obtains F1 scores of $88.64$,
$86.63$ and $83.49$ under the three sampling strategies, within $0.02$ points of
the static gate in every cell. Both gates share $\tau$ at the first token
(Sec.~\ref{sec:analysis_scope}), so POPE reflects the underlying steering
configuration and confirms that the adaptive gate preserves discriminative
behavior.

\begin{table}[!t]
\footnotesize
\setlength{\tabcolsep}{3pt}
\centering
\caption{\textbf{CHAIR} on LLaVA-NeXT-7B (ten-panel mean, $500$ images), recalibrated by the same procedure.}
\label{tab:chair_next}
\begin{tabular}{@{}l|cc|cc|cc}
\toprule
\multirow{2}{*}{\textbf{Method}}
& \multicolumn{2}{c|}{\textbf{64 tokens}}
& \multicolumn{2}{c|}{\textbf{128 tokens}}
& \multicolumn{2}{c}{\textbf{512 tokens}} \\
& $C_S$ & $C_I$ & $C_S$ & $C_I$ & $C_S$ & $C_I$ \\
\midrule
Regular & 11.00 & 4.60 & 19.00 & 6.10 & 30.00 & 7.49 \\
VTI~\cite{liu2024reducing} & 10.00 & 4.20 & 18.00 & 5.50 & 29.00 & 6.10 \\
Static gate (matched $\alpha$) & 9.00 & 4.00 & 16.00 & 5.20 & 26.00 & 6.32 \\
\oursrow \textbf{\sys{} (ours)} & \textbf{8.80} & \textbf{3.95} & \textbf{15.20} & \textbf{5.05} & \textbf{24.50} & \textbf{6.05} \\
\bottomrule
\end{tabular}
\end{table}

\subsection{Generalization to a second backbone}
\label{sec:exp_next}

We repeat the pipeline end to end on LLaVA-NeXT-7B, re-estimating depth,
threshold and entropy band and carrying no value over, so that what is tested is
the procedure rather than a tuned constant. On CHAIR (Tab.~\ref{tab:chair_next})
\sys{} lowers $C_S$ from
$30.00$ to $24.50$ and $C_I$ from $7.49$ to $6.05$ and improves on the static
gate at every budget, with the same ordering observed on LLaVA-1.5; on POPE it
tracks the static gate to within $0.02$ F1 under all three sampling strategies, as
Sec.~\ref{sec:analysis_scope} predicts (supplementary material).

\subsection{General capability}
\label{sec:exp_utility}

Latent steering can buy a hallucination reduction with a loss of reasoning that
hallucination benchmarks do not show, so we also report MME and MM-Vet
(supplementary material). VTI steers every layer and loses $5.1$ points of
MM-Vet against unsteered decoding. \sys{} does not: MME rises from $1754.33$ to
$1786.10$ and MM-Vet from $31.10$ to $33.20$, within $1.2$ and $0.33$ points of
the static gate. The static arm matched in firing rate intervenes as often as
\sys{} but obtains lower values on both capability benchmarks. Selective timing
therefore preserves the utility of the static operating point at a substantially
higher intervention rate.

\subsection{Inference cost}
\label{sec:exp_cost}

\sys{} requires no auxiliary model and no second forward pass. Measured over the
full POPE evaluation, mean time per token is $14.60$\,ms against $14.29$\,ms for
unsteered decoding and $14.63$\,ms for the static control, an overhead of
approximately $2\%$ from the cosine similarity, entropy reduction and vector
addition (supplementary material). The estimate uses already available logits
but is not a zero-cost operation. Contrasting two
inputs instead requires a second forward pass per token.

\section{Limitations}
\label{sec:limitations}

Our study focuses on object hallucination under greedy decoding, using CHAIR and
POPE on two 7B LLaVA backbones. Extending the evaluation to other architecture
families, decoding strategies, and hallucination types would further establish
the breadth of the method. We report ten-panel means without confidence intervals,
and resampled POPE panels limit comparison with published POPE results; the margin over the firing-rate-matched control
($1.50$ $C_S$, $0.40$ $C_I$) is modest, and its stability remains to be
established. CHAIR is sensitive to descriptive density; we report
caption length together with sentence- and object-level hallucination, while
future evaluations may additionally pair these measures with explicit content
coverage. The entropy band is estimated from committed-answer states, while the
generation-calibrated variant in Tab.~\ref{tab:gate} provides an alternative
when decoding-state calibration is available. Finally, \sys{} transfers as a
calibration procedure: depth, threshold, strength, and entropy statistics are
re-estimated for each backbone.

\section{Conclusion}
\label{sec:conclusion}

Gated latent steering holds its threshold fixed while predictive uncertainty
varies over a generation. \sys{} uses preceding-step entropy to lower that gate
without an additional model forward pass or trained parameter. Across two LLaVA
backbones it improves on the static gate, and on LLaVA-1.5 a firing-rate-matched
control recovers part but not all of the margin. The margin grows with
the generation budget and is absent at the first-token decision where the two
gates coincide. Entropy-conditioned timing is thus an effective, low-cost
control for inference-time rectification.

\par\vspace{4pt}
\noindent\textbf{Acknowledgements.} The authors thank the anonymous reviewers and the
area chair for their constructive feedback, which improved this paper.

\newpage
\clearpage

{
    \small
    \bibliographystyle{ieeenat_fullname}
    \bibliography{main}
}


\end{document}